\documentclass{article}
\usepackage[preprint]{neurips_2019}

\usepackage[utf8]{inputenc}
\usepackage[T1]{fontenc}
\usepackage{mathtools, amssymb}
\usepackage{microtype}
\usepackage{hyperref}
\usepackage{natbib}

\usepackage{pgfplots}
\usepackage{pgfplotstable}
\pgfplotsset{compat=1.13}
\pgfplotstableset{%
    col sep=comma,
    columns={n,0.5-robustness, 0.8-robustness,0.95-robustness,0.99-robustness,1-robustness},
    fixed zerofill,
    fixed,
    precision=3,
    columns/n/.style={
        int detect,column type=r,column name=$n$,
    },
    empty cells with={--}, 
    every head row/.style={before row=\toprule,after row=\midrule},
    every last row/.style={after row=\bottomrule}
}

\usepackage{braket}
\usepackage{comment}

\usepackage{booktabs}
\usepackage{tabularx}

\usepackage{float}
\usepackage{graphicx}

\usepackage{listings}
\usepackage{color}

\definecolor{mygreen}{rgb}{0,0.6,0}
\definecolor{mygray}{rgb}{0.5,0.5,0.5}
\definecolor{mymauve}{rgb}{0.58,0,0.82}

\newcommand{\dmnamesing}{RPENN}
\newcommand{\dmnameplur}{RPENNs}

\newcommand{\AEMfgsm}{FGSM}

\newcommand{\AEMbimlinf}{$\text{BIM}_{\ell_\infty}$}
\newcommand{\AEMdfa}{DFA}
\newcommand{\AEMsma}{SMA}
\newcommand{\AEMlbfgs}{L-BFGS}
\newcommand{\AEMcwltwo}{$\text{CW}_{\ell_2}$}

\newcommand{\dnnvar}{h}
\newcommand{\dmvar}{d}
\newcommand{\reldev}{\lambda}
\newcommand{\robulevel}{q}
\newcommand{\esbsize}{m} 
\newcommand{\ninstalls}{n}
\newcommand{\dnnplusdm}{\dnnvar + \dmvar} 
\newcommand{\seedvar}{\sigma}

\newcommand{\seqspace}{\mathcal{S}}

\newcommand{\testseq}{S}
\newcommand{\nummatches}[2]{\mathbb{M}\left({#1}, {#2}\right)}
\newcommand{\efficacyvar}{\mathcal{E}}
\newcommand{\qualityvar}{\mathcal{Q}}
\newcommand{\robuvar}{\mathcal{R}}

\DeclarePairedDelimiterX{\card}[1]{\lvert}{\rvert}{#1}
\DeclarePairedDelimiterX{\norm}[1]{\lVert}{\rVert}{#1}
\DeclarePairedDelimiter{\ceil}{\lceil}{\rceil}
\DeclarePairedDelimiter{\floor}{\lfloor}{\rfloor}

\newcommand{\reals}{\mathbb{R}}

\newcommand{\normal}[2]{\mathcal{N}\left(#1, #2\right)}
\newcommand{\defeq}{\vcentcolon=}

\title{An Empirical Investigation of \\Randomized Defenses against Adversarial Attacks}

\author{%
    Yannik Potdevin \\
    Kiel University \\
    Germany \\
    \texttt{ypo@informatik.uni-kiel.de} \\
    \And
    Dirk Nowotka \\
    Kiel University \\
    Germany \\
    \texttt{dn@informatik.uni-kiel.de} \\
    \And
    Vijay Ganesh \\
    Waterloo University \\
    Canada \\
    \texttt{vijay.ganesh@uwaterloo.ca} \\
}

\begin{document}
\maketitle
\begin{abstract}
In recent years, Deep Neural Networks (DNNs) have had a dramatic impact on a variety of problems that were long considered very difficult, e.\,g., image classification and automatic language translation to name just a few. The accuracy of modern DNNs in classification tasks is remarkable indeed. At the same time, attackers have devised powerful methods to construct specially-crafted malicious inputs (often referred to as \emph{adversarial examples}) that can trick DNNs into mis-classifying them. What is worse is that despite the many defense mechanisms proposed to protect DNNs against adversarial attacks, attackers are often able to circumvent these defenses, rendering them useless. This state of affairs is extremely worrying, especially since machine learning systems get adopted at scale.

In this paper, we propose a scientific evaluation methodology aimed at assessing the quality, efficacy, robustness and efficiency of randomized defenses to protect DNNs against adversarial examples. Using this methodology, we evaluate a variety of defense mechanisms. In addition, we also propose a defense mechanism we call Randomly Perturbed Ensemble Neural Networks (\dmnameplur). We provide a thorough and comprehensive evaluation of the considered defense mechanisms against a white-box attacker model, six different adversarial attack methods and using the ILSVRC2012 validation data set.
\end{abstract}

\section{Introduction}
Machine learning algorithms – especially Deep Neural Networks (DNNs) – have received considerable attention in recent years thanks to their ability in addressing long-standing difficult problems in diverse areas such as computer vision \cite{DBLP:conf/nips/KrizhevskySH12}, speech recognition \cite{DBLP:journals/spm/X12a}, games~\cite{DBLP:journals/nature/SilverHMGSDSAPL16}, and autonomous driving~\cite{DBLP:journals/corr/BojarskiTDFFGJM16}, to name just a few. Unfortunately, while neural network classifiers seem remarkably successful when applied within a benign environment, they tend to fail spectacularly when confronted with so called \emph{adversarial examples} (AEs). This was powerfully demonstrated in the work of Szegedy et al. \cite{DBLP:journals/corr/SzegedyZSBEGF13}, who were the first to show that neural network classifiers were vulnerable to adversarial inputs.\footnote{
In this paper, we interchangeably use the terms inputs and examples.
}

Informally, given an input $x$ for a neural network $\dnnvar$, an adversarial example is an input $x'$ that is indistinguishable from $x$ for a human observer, but causes $\dnnvar$ to behave in an undesirable way, i.\,e., different from how $\dnnvar$ behaves on $x$. (In Section \ref{sect:background}, we precisely define many terms used informally here.)

In the years since Szegedy et al.’s work was published, many methods to generate adversarial examples have been developed, often enabling the attacker to choose the particular way the targeted network should misbehave \cite{DBLP:conf/sp/Carlini017,DBLP:journals/corr/GoodfellowSS14,DBLP:journals/corr/KurakinGB16,DBLP:conf/cvpr/Moosavi-Dezfooli16,DBLP:conf/eurosp/PapernotMJFCS16,DBLP:journals/corr/SzegedyZSBEGF13}. Also, several methods to defend against  \cite{DBLP:journals/corr/GuR14,DBLP:conf/cvpr/LiaoLDPH018,DBLP:conf/sp/PapernotM0JS16,DBLP:journals/corr/TramerKPBM17,DBLP:conf/ndss/Xu0Q18}, and methods to detect adversarial inputs, have been developed. For an overview see the paper \cite{DBLP:conf/ccs/Carlini017} by Carlini and Wagner. However, many of these methods have been shown to be less effective than claimed, for example, by Carlini and Wagner in \cite{DBLP:conf/ccs/Carlini017,DBLP:journals/corr/abs-1711-08478,DBLP:conf/sp/Carlini017}.

As Goodfellow et al. point out in \cite{DBLP:journals/cacm/GoodfellowMP18}, it is of great importance to make systems incorporating machine learning software robust against adversarial examples, especially since it has been shown by Kurakin et al. in \cite{DBLP:journals/corr/KurakinGB16} and by Brown et al. in \cite{DBLP:journals/corr/abs-1712-09665} that adversarial examples can be harmful in the physical world. For example, regarding the safety of autonomous driving, Eykholt et al. have demonstrated in \cite{DBLP:conf/cvpr/EykholtEF0RXPKS18} that real traffic signs may be practically manipulated to be misinterpreted by image classification neural networks.

\vspace{0.2cm}
\noindent{\textbf{Problem Statement:}} \emph{To what extent are defense mechanisms (DMs), which utilize randomization, able to defend DNNs against adversarial examples?} In this paper we present a comprehensive methodology to empirically evaluate (randomized) defense mechanisms. Using that methodology, we answer the former question exemplarily, by investigating defense mechanisms which utilize randomization. Two of them are proposed by Gu and Rigazio in \cite{DBLP:journals/corr/GuR14}, one of them is an extension of their work which we will present and the last one is \dmnamesing\ – a novel defense mechanism we introduce in this work.

\vspace{0.2cm}
\noindent{\textbf{Contributions:}} In this paper, we make the following contributions:
\begin{enumerate}
    \item {\bf Evaluation Methodology for NN Defense Mechanisms:} Our most important contribution is an evaluation methodology that enables us to assess defense mechanisms over a set metrics and associated properties such as \emph{efficacy}, \emph{quality}, \emph{robustness} and \emph{efficiency} (see Section \ref{sect:properties}). We compare several DMs along these metrics. More specifically, using the Taxonomy given by Papernot et al. \cite{DBLP:journals/corr/PapernotMSW16}, we define a precise threat model (see Section \ref{sect:background}). Based on that model, we reason about properties which enable a defense mechanism be {\it effective} against such attackers. We then derive natural metrics to measure and compare those properties across defense mechanisms (refer to Section \ref{sect:properties}). To make our results robust, we evaluate on the real world sized VGG19~\cite{DBLP:journals/corr/SimonyanZ14a} deep neural network, the large and popular validation data set ILSVRC2012~\cite{deng2009imagenet}\footnote{%
        In case the reader wonders why we use the data set from 2012: It was introduced in that year and reused up to 2017.
    }
    and six different methods \cite{DBLP:journals/corr/GoodfellowSS14, DBLP:journals/corr/KurakinGB16, DBLP:conf/cvpr/Moosavi-Dezfooli16, DBLP:conf/eurosp/PapernotMJFCS16} to generate adversarial examples (the L-BFGS attack \cite{DBLP:journals/corr/SzegedyZSBEGF13} and the Carlini Wagner $\ell_2$ attack \cite{DBLP:conf/sp/Carlini017} are among them).

    Further, in order to make our experiments comprehensible and to enable other researchers to repeat them, we carefully describe the execution of our evaluation by covering preprocessing, adversarial example generation, metric estimation, and the choice of DM parameters. We also publish our implementation of the considered defense mechanisms and the benchmark utilities at GitHub.\footnote{
        See \url{https://github.com/ypotdevin/randomized-defenses}.
    } Finally, our methodology conforms with the suggestions given by Carlini and Wagner in \cite{DBLP:conf/ccs/Carlini017}.
    \vspace{0.2cm}

    \item {\bf Randomly Perturbed Ensemble Neural Networks:} We also propose a randomized defense mechanism, called Randomly Perturbed Ensemble Neural Networks (\dmnameplur), that generates from a trained neural network $\dnnvar$ several randomized variations – a randomized ensemble – which are then used for inference instead of the original network. For a detailed description of \dmnamesing, see Section \ref{sect:dm}.
    \vspace{0.2cm}

   \item {\bf Evaluation and extensions of DMs proposed by Gu and Rigazio:} Among the first mechanisms proposed to defend against adversarial examples were the L1 and L* defense mechanism by Gu and Rigazio~\cite{DBLP:journals/corr/GuR14}. The key feature of L1 is to inject noise into the input of a DNN, before inferring it. In L* the noise is additionally injected into the hidden layer of a DNN, before inferring the input. As a natural extension of Gu and Rigazio’s work, we introduce L+, were the noise is only injected into the hidden layers before inferring the input. We evaluate the three DMs way more extensively than was done previously in the literature.

   \vspace{0.2cm}
   \item {\bf Important Findings:} In addition to developing a robust methodology for evaluating randomized DMs, we make several surprising findings. As mentioned above, we evaluated 4 different DMs, namely, L1, L*, L+, and RPENN. Of these, L1 is the simplest, since it only perturbs the input layer. To our surprise, L1 performed best against the other DMs on the properties we tested, namely, efficacy, quality, robustness, and efficiency.
\end{enumerate}

Besides the above mentioned properties, an often overlooked aspect of DMs is whether they are cost-effective to deploy, a property we call deployability. While this property is hard to define mathematically, it is nonetheless very important to consider in practical settings. For example, in traditional computer security context, many DMs can be easily deployed via a software patching mechanism over the internet. Fortunately, all DMs discussed in this paper also share this »ease of deployment« property, and any deployed DNN can be easily patched with the DMs we study. (We shall not discuss this property any further in this paper.)

\vspace{0.2cm}
\noindent{\textbf{Structure of the Paper:}} The rest of this paper is structured as follows: In Section \ref{sect:background}, we provide definitions for concepts such as adversarial examples/attacks, defense mechanism (DM), and a threat model (Subsection \ref{sect:threat-model}). In Section \ref{sect:metrics}, we define the notion and properties of {\it good defense mechanisms}. In Section \ref{sect:dm}, we describe the various DMs considered in this paper, including the ones we introduced, namely, RPENN and L+. In Section \ref{sect:metho}, we describe our experimental methodology, followed by results in Section \ref{sect:results}. In Section \ref{sect:related}, we compare our work to previous work. In Section \ref{sect:validity}, we discuss possible weaknesses of our experimental approach and steps we took to mitigate them. Finally, we conclude in Section \ref{sect:conc}.

\section{Preliminaries}
\label{sect:background}
In this section, we define many terms used throughout the paper. Wherever appropriate, we make definitions mathematically precise. Given that the field of ML security is still in its infancy, many terms are only informally defined.

\paragraph{Deep Neural Networks} For the sake of completeness, we provide a brief definition of deep neural networks (DNNs), specifically focusing on those that perform image classification tasks, i.\,e., \emph{classifiers}. More precisely, we model images to be real valued vectors of a domain $D \subseteq \reals^i$, class labels to be integers in $\Set{1, \dots, j}$, where $j$ is the number of different classes considered, and a classifier to be a function $\dnnvar \colon D \to \Set{1, \dots, j}$.

\paragraph{Adversarial Examples:} Given an image $x \in D$ which is correctly classified by a classifier $\dnnvar$, an \emph{adversarial example} is an image $x' \in D$ such that $\dnnvar(x) \neq h(x')$ and $\mathrm{sim}(x, x')$ is low, where $\mathrm{sim} \colon \reals^i \times \reals^i \to \reals_{\geq 0}$ is a suitable\footnote{
    Often, for simplicity, the Euclidean distance, the Chebyshev distance or the $\ell_0$-“norm” are chosen as a metric. It is debatable whether these choices truly capture similarity between images. However, in practice they mostly yield satisfying results.
} metric capturing the similarity between the images $x$ and $x'$.

As an example, let $x \in D$ be an image correctly classified by $\dnnvar$ as a Panda. Then an adversarial example is an image $x' \in D$, where only some pixels differ slightly from the respective pixels in $x$ (a human would still classify $x'$ as a Panda, probably one would not even see a difference between $x$ and $x'$), but $\dnnvar(x')$ is the class for images of Gibbons.

\paragraph{Defense Mechanism:}
By defense mechanisms we mean a function (or from the practical point of view: a computer program) $\dmvar$ taking a classifier $\dnnvar$ as input and yielding a classifier as output $\dmvar(\dnnvar)$, which is of the same domain and codomain as the input classifier.
Instead of $\dmvar(\dnnvar)$, we use the notation $\dnnplusdm$ throughout this paper.
We may parameterize a \emph{randomized} defense mechanism, like \dmnamesing, additionally by its random seed $\seedvar$ denoted by $d_{\seedvar}$. When the context is clear or the seed variable is not essential for the presentation, we omit the $\seedvar$ and just write $\dmvar$.

\paragraph{Installation:}
By the term \emph{installation} of a classifier $\dnnvar$ (which specifically includes classifiers protected by a DM), we refer to copy of a computer program realizing $\dnnvar$ on a device. This term has the same connotation as the \emph{installation} of an operating system on a desktop computer.
We point out that realizing a classifier $\dnnvar$ may include providing additional information to it, which are highly depending on (the state of) the ambient device (like a random seed for example).
This way one installation realizing $\dnnvar$ might show different input output behavior than another installation realizing $\dnnvar$.

\subsection*{Threat Model}
\label{sect:threat-model}
Following the taxonomy given in Papernot et al. \cite{DBLP:journals/corr/PapernotMSW16}, a threat or attacker model suitable for ML systems consists of three main components:
\begin{enumerate}
    \item Attack surface (the parts of the ML system that an attacker can access and manipulate)
    \item Adversarial capabilities (the actions and information an attacker may use)
    \item Adversarial goals (what an adversary tries to achieve)
\end{enumerate}
We use this relatively informal threat model, walking through the three main components of it, to describe both the capabilities and limitations of an attacker.

\paragraph{Attack Surface:} The parts of the attack surface of a machine learning system which the attacker may attack in the inference phase are: the physical input domain (objects, actions), the digital representation (bits) of the relevant subset of the physical domain, the hardware involved in generating that digital representation (sensors, cameras, IO hardware), the machine learning model, its input features and the procedure of preprocessing the digital representation to those features.

\paragraph{Adversarial Capabilities:} The attacker is capable of manipulating the input features of the ML systems. The attacker may achieve this by directly accessing them, or by modifying any previous stage – the preprocessing procedure, the digital representation, the sensors or the physical objects. Additionally, the attacker has white box read access to the machine learning model itself, including its architecture and its weights. The attacker or adversary may not attack any of the following: any part or the entirety of the training process, the application of the machine learning model and its outputs, the output analysis, the actions taken based on the outputs,
and most importantly: the seeds used to realize the randomization.

\paragraph{Adversarial Goals:} The goal of the adversary is to attack the integrity and availability of the ML system, by forcing the ML system to (mis)classify in his favor. For example, the attacker may bypass biometric access control systems or he may force autonomous cars to skip stop signs. However, the attacker may not violate confidentiality or privacy properties of the ML system under attack.

\section{Properties of Good Defense Mechanisms}
\label{sect:properties}
Here we articulate a set of properties that we believe all \emph{good} defense mechanisms must possess. To the best of our knowledge, we are not aware of any other effort at articulating explicitly such properties of ML defense mechanisms. We also characterize how these properties can be evaluated experimentally.%
\footnote{
    It has to be noted that since ML security research is still in its infancy, we are far from mathematically proving that any DMs possess said good properties. We are, at this stage, limited to performing empirical evaluation of proposed DMs within the context of mathematically defined metrics, chosen data sets, fixed attack models, as well as the kinds of DNNs being protected. Hence, it goes without saying that all empirical conclusions are contingent on these choices. Having said that, we still take great pains here to define these properties as crisply as possible such that it enables us to make precise measurements, analyses, and appropriate comparisons.
}

\paragraph{Efficacy:} Informally, we say that a DM $\dmvar$ is \emph{effective}, if the accuracy of $\dnnplusdm$ on adversarial inputs is higher than the accuracy of $\dnnvar$.
The metric we use to measure efficacy of a DM $\dmvar$ is the natural one of taking the ratio of the number of correctly classified adversarial inputs over the total number of adversarial inputs.
Ideally, we want our DMs to simultaneously protect the input DNN from adversarial inputs as well as maintaining (or improving) its accuracy on benign inputs.

\paragraph{Quality:} Informally, we say that a defense mechanism $\dmvar$ is of high quality, if $\dnnplusdm$ has similar or better accuracy than $\dnnvar$ on benign inputs.

\paragraph{Robustness:} Informally, we say that a defense mechanism $\dmvar$ is robust, if any single adversarial example is only misclassified by a small fraction of the installations of $\dnnplusdm$. This is similar to the diversity property of certain randomization-based defense mechanisms used in the context of system security, e.\,g., Address Space Layout Randomization (ASLR).

\paragraph{Efficiency:} Informally, we say that a defense mechanism $\dmvar$ is efficient, if applying $\dnnplusdm$ introduces just constant overhead regarding running time and space requirements, compared to applying just $\dnnvar$.

    \subsection*{Metrics to Quantify Properties of Defense Mechanisms}
    \label{sect:metrics}
    Here we mathematically define the metrics used to quantify the above-mentioned properties of defense mechanisms. Let $\seqspace \defeq (D \times \Set{1, \dots, j})^{*}$ denote the set of finite sequences of pairs, where the left components are images and the right components are class labels.
    Let $\testseq \in \seqspace$ be a sequence of images and their corresponding true labels and $\dnnvar$ be a classifier. We denote by $\dnnvar(\testseq)$ the sequence of labels $\dnnvar$ predicts for the images in $\testseq$. By $\mathcal{A}_{\dnnvar}(\testseq) \in \seqspace$ we denote a sequence of adversarial examples an attacker $\mathcal{A}$ derived from $\testseq$ to attack $\dnnvar$. The class labels in $\mathcal{A}_{\dnnvar}(\testseq)$ are $\mathcal{A}$’s target labels, those labels which are different from the ones in $\testseq$ and which $\mathcal{A}$ desires to match with $\dnnvar(\mathcal{A}_{\dnnvar}(\testseq))$.
    For any label sequence $L$
    we define by $\nummatches{L}{\testseq} \defeq \card*{\Set{ i | L_i = (\testseq_i)_2 }}$ the number of matching labels of $L$ and $\testseq$ (note that the subscript 2 in $(\testseq_i)_2$ indicates the right component of the pair $\testseq_i$).

    Now we are able to define metrics to compare defense mechanisms regarding several properties.

        \paragraph{Efficacy Metric:}
        We evaluate the ability of a defense mechanism $\dmvar$ to block adversarial examples with respect to 1) a classifier to protect $\dnnvar$, 2) an adversarial example generation method $\mathcal{A}$ and 3) a test sequence $\testseq$ which $\mathcal{A}$ used to derive adversarial examples from.
        Having fixed these, we measure $\dmvar$’s efficacy by
        \[
            \efficacyvar_{\dnnvar, \mathcal{A}, S}(\dmvar) \defeq
                \frac{ \nummatches{(\dnnplusdm)(\mathcal{A}_{\dnnvar}(\testseq))}{\testseq} }
                     { \card{\mathcal{A}_{\dnnvar}(\testseq)} }.
        \]
        That is the amount of adversarial examples classified correctly (regarding their ground truth) divided by the total amount of adversarial examples. One could also say this is the Top-1 accuracy of $\dnnplusdm$ on the adversarial examples sequence.

        \paragraph{Quality Metric:}
        Using the same prerequisites as for the efficacy metric, disregarding the presence of an attacker, we measure $\dmvar$’s quality by
        \[
            \qualityvar_{\dnnvar, S}(\dmvar) \defeq
                \frac{ \nummatches{(\dnnplusdm)(\testseq)}{\testseq}}
                     {\lvert \testseq \rvert}.
        \]
        This is simply the Top-1 accuracy of $\dnnplusdm$ on the test sequence $\testseq$.

        \paragraph{Robustness Metric:}
        Adding to the prerequisites needed to define the efficacy metric, let $\seedvar_1, \seedvar_2, \dots$ be a sequence of random seeds and $\dmvar_{\seedvar_1}, \dmvar_{\seedvar_2}, \dots$ be a sequence of installations of a randomized defense mechanism $d$.
        The robustness of a defense mechanism is measured, with respect to a number of installations $n$ and a robustness level $\robulevel \in \reals, 0 < \robulevel \leq 1$, by
        \begin{align*}
            \MoveEqLeft \robuvar_{\dnnvar, \mathcal{A}, S}(d, \robulevel, \ninstalls) \defeq \\
                &  \frac{ \card*{\Set{ A_i |
                                                \nummatches{ \langle L_{i, 1}, \dots, L_{i, n} \rangle }
                                                           { \langle A_i, \dots, A_i \rangle }
                                                \geq
                                                \floor{\robulevel \cdot \ninstalls}
                                           }
                                      }
                               }
                     { \card{\mathcal{A}_{\dnnvar}(\testseq)} },
        \end{align*}
        where $A$ is a shorthand for $\mathcal{A}_{\dnnvar}(\testseq)$
        and $L \defeq \langle (\dnnplusdm_{\seedvar_1})(A)^T, \dots, (\dnnplusdm_{\seedvar_n})(A)^T \rangle$ is a label matrix combining as columns the predicted class label sequences of the $n$ installations (with their seeds) for the adversarial inputs $A$. In the enumerator only those adversarial inputs are counted, which are misclassified as intended by at least $\floor{\robulevel \cdot \ninstalls}$ out of $\ninstalls$ installations of $d$. Thus the robustness metric estimates the attacker’s chance of corrupting at least a $\robulevel$-fraction of $n$ installations.
        For any robustness level $\robulevel$ it is desirable that $\robuvar_{\dnnvar, \mathcal{A}, S}(d, \robulevel, \ninstalls)$ approaches 0 considerably fast, as $n$ increases.

        When it is clear from the context, we omit the indices $\dnnvar$, $\mathcal{A}$, $\testseq$ from the metrics (and regarding $\robuvar$ even $\dmvar$) to give a more crisp presentation.

\section{Randomized Defense Mechanisms}
\label{sect:dm}
In this section we describe the noise injection defense mechanisms proposed by Gu and Rigazio in \cite{DBLP:journals/corr/GuR14}, which they denoted »L1« and »L*«. We also introduce a natural extension of their work, which we will call »L+«. This section ends with the introduction our defense mechanism: \dmnamesing.

\subsection*{The L1 defense mechanism}
Since Gu and Rigazio did not publish a reference implementation, we had to reimplement their mechanisms in order to assess them. In \cite[Section~3.1]{DBLP:journals/corr/GuR14}, Gu and Rigazio describe their two defense mechanisms only roughly, stating that the noise is \emph{added to layers} – not clarifying if the noise is added to the weights of the layers, to the input of their activation functions, to the output of their activation functions, or any combination of those. We decided to interpret Gu and Rigazio’s description as adding noise to the weights, because it is easy to implement and also more comparable to our defense mechanism (which we will explain below). We make an exception to that rule regarding the input layer, since that layer has no weights (and no activation function). In that case, we add the noise directly to the input.

The core idea behind the L1 defense mechanism is to perturb the input of a DNN with noise, hoping that this way the adversarial manipulation of the input is neutralized and the ground truth restored. The amount of noise which will be injected is controlled by the parameter $\sigma \in \reals_{> 0}$. In more detail:
\begin{enumerate}
    \item
    Read the input $x = \langle x_1, \dots, x_i \rangle$ of a DNN $\dnnvar$.
    \item
    Draw independently $i$ random samples $\epsilon_1, \dots, \epsilon_i$ from the Gaussian distribution $\normal{0}{\sigma^2}$.
    \item
    With $x' \defeq \langle x_1 + \epsilon_1, \dots, x_i + \epsilon_i \rangle$, return $y = \dnnvar(x')$.
\end{enumerate}

\subsection*{The L* defense mechanism}
In the L* defense mechanism the idea of noise injection is taken to the next level, by perturbing additionally the weights of the DNN. In more detail:
\begin{enumerate}
    \item
    Read the input $x = \langle x_1, \dots, x_i \rangle$ of a DNN $\dnnvar$, with weights $w_1, w_2, \dots, w_k$.
    \item
    Draw independently $i + k$ random samples $$\epsilon_1, \dots, \epsilon_i, \epsilon_{i + 1}, \dots, \epsilon_{i + k}$$ from the Gaussian distribution $\normal{0}{\sigma^2}$.
    \item
    Replace the weights $w_1, w_2, \dots, w_k$ of $\dnnvar$ by $$w_1 + \epsilon_{i + 1}, w_2 + \epsilon_{i + 2}, \dots, w_k + \epsilon_{i + k}.$$
    \item
    With $x' \defeq \langle x_1 + \epsilon_1, \dots, x_i + \epsilon_i \rangle$, return $y = \dnnvar(x')$.
\end{enumerate}
Note that Gu and Rigazio only expose the parameter $\sigma$ in their description of L1 and L*. This is why we assume that the noise aiming at the input is drawn from the same distribution as the noise aiming at the weights of the hidden layers is drawn from.

\subsection*{The L+ defense mechanism}
To see what effect perturbing the weights of the hidden layers without perturbing the input would have, we also implemented a DM we call L+, which is essentially L* minus L1. In more detail:
\begin{enumerate}
    \item
    Read the input $x$ of a DNN $\dnnvar$, with weights $w_1, w_2, \dots, w_k$.
    \item
    Draw independently $k$ random samples $\epsilon_{1}, \dots, \epsilon_{k}$ from the Gaussian distribution $\normal{0}{\sigma^2}$.
    \item
    Replace the weights $w_1, w_2, \dots, w_k$ of $\dnnvar$ by $$w_1 + \epsilon_{1}, w_2 + \epsilon_{2}, \dots, w_k + \epsilon_{k}.$$
    \item
    Return $y = \dnnvar(x)$.
\end{enumerate}
One could also understand L* as L1 plus L+ which is, in our opinion, a more suitable point of view, since we rate editing the input of a DNN to be conceptually different from editing the DNN itself. When Gu and Rigazio introduced L1 and L*, we argue that they presented two different concepts, one is given by L1 and the other is partially hidden in L*. We make that hidden concept explicit by presenting L+.

\subsection*{The RPENN defense mechanism}
In this section, we describe our own defense mechanism to protect deep neural networks from adversarial examples: Randomly Perturbed Ensemble Neural Networks (RPENNs). Informally, we take a trained DNN $\dnnvar$, generate $\esbsize$ copies of it, and randomly perturb the weights of each copy. We then compose these DNNs into an ensemble via majority voting.  

The key insight that led us to developing our defense mechanism is the following:
If adversarial examples are hard to distinguish from their benign twins for the human viewer, how come that DNNs come to high confidence mis-classifications? Is it because there are very sensitive (combinations of) units in the network reacting to the tiniest change of their input? Do adversarial example generation methods rely on the very setting of these sensitive units to exploit them? If so, maybe it would help to reduce the sensitivity of those units by manipulating their weights a little and therefore changing the setting the attacker relied on.

Guided by the assumption stated above, we developed a defense mechanism which takes a trained DNN $\dnnvar$ (ready for production use), copies its weights (and biases) and generates, based on the copied weights, randomized variations $\dnnvar_1, \dnnvar_2, \dots \dnnvar_{\esbsize}$ which are similar to $\dnnvar$, but not equal (with a high probability). Then the inputs predetermined for $\dnnvar$ are fed to $\dnnvar_1, \dnnvar_2, \dots \dnnvar_{\esbsize}$ instead, processed by them, and their respective outputs are combined to a new final result.
In more detail:
\begin{enumerate}
    \item
    Read the input $x$ of a DNN $\dnnvar$, with weights $w_1, w_2, \dots, w_k$.
    \item
    Draw independently new weights
    \[
    \begin{matrix}
        w^{1}_{1} \sim \normal{ w_1}{(\lambda \cdot w_1)^2}
        & \dots
        & w^{1}_{k} \sim \normal{ w_k}{(\lambda \cdot w_k)^2} \\
        w^{2}_{1} \sim \normal{ w_1}{(\lambda \cdot w_1)^2}
        & \dots
        & w^{2}_{k} \sim \normal{ w_k}{(\lambda \cdot w_k)^2} \\
        \vdots & \ddots & \vdots \\
        w^{\esbsize}_{1} \sim \normal{ w_1}{(\lambda \cdot w_1)^2}
        & \dots
        & w^{\esbsize}_{k} \sim \normal{ w_k}{(\lambda \cdot w_k)^2},
    \end{matrix}
    \]
    where $\lambda \in \mathbb{R}_{>0}$ is a user defined parameter controlling the amount of deviation from the original weight $w_{i}$,
    called relative deviation,
    and therefore $\normal{w_{i}}{(\lambda \cdot w_{i})^2}$ is a normal distribution with mean $w_i$ and variance $(\lambda \cdot w_{i})^2$.
    \item
    Generate $\esbsize$ new networks $\dnnvar_1, \dnnvar_2, \dots, \dnnvar_{\esbsize}$,
    the variations of $\dnnvar$,
    where the weights of each variation $\dnnvar_i$ are $w^{i}_{1}, w^{i}_{2}, \dots, w^{i}_{k}$.
    \item
    Feed $x$ to $\dnnvar_1, \dnnvar_2, \dots, \dnnvar_{\esbsize}$,
    creating the outputs $$\dnnvar_1(x), \dnnvar_2(x), \dots, \dnnvar_{\esbsize}(x).$$
    \item
    Combine $\dnnvar_1(x), \dnnvar_2(x), \dots, \dnnvar_{\esbsize}(x)$ to a single output $y$,
    for example by averaging or by picking the majority vote.
    \item
    Return $y$.
\end{enumerate}
\label{passage:key-params}
We point out that two parameters drastically affect the behavior of RPENN, namely the parameter $\lambda$, determining the relative deviation and the parameter $\esbsize$, determining the number of variations to generate – the ensemble size. We refer to Section \ref{sect:results} for how these two parameters influence the \dmnamesing’s quality, efficacy, robustness and efficiency.

Note that choosing $\esbsize = 1$ disables the ensemble, and reveals one main difference between RPENN and L+: The weight-individual Gaussian distributions used to redraw the weights of the DNN variant. Utilizing individual distributions like we did ensures that each weight is – relatively – perturbed by the same amount, disregarding the fact that some weights might be orders of magnitude larger than other weights. Using the same level of noise for all weights might on one hand substantially perturb small weights, while on the other hand it might have negligible effects on large weights.

Briefly speaking: One important ingredient of RPENN is the usage of individual distributions and the other one is the usage of an ensemble to leverage self-correcting behavior.
\section{Experimental Methodology}
\label{sect:metho}
In this section we justify our choice of the ImageNet data set and the VGG19 convolutional network, we describe what preprocessing we applied, how we generated the considered adversarial examples, how we selected the parameters of the defense mechanisms and how we estimated\footnote{
    Since the defense mechanisms are randomized, we cannot determine their metrics with certainty; we can only estimate them.
} the quality, the efficacy and the robustness of the assessed defense mechanisms.

\paragraph{Selection of Benchmark Data and Model}
To get realistic and relevant results, we chose to evaluate the defense mechanisms using the validation data set belonging to the ImageNet \cite{deng2009imagenet} Large Scale Visual Recognition Challenge 2012 (ILSVRC 2012). That set contains 50000 color images and each picture has a resolution of $224 \times 224$, meaning that its resolution is significantly higher than, for example, the resolution of the $28 \times 28$ gray scale images belonging to the MNIST \cite{lecun1998gradient} data set. This choice is supported by Carlini and Wagner in \cite{DBLP:conf/ccs/Carlini017}, where they suggest evaluating defense mechanisms not just against MNIST, but against CIFAR-10 or, even better, against ImageNet.

The DNN we used to benchmark the defense mechanisms is the VGG19 \cite{DBLP:journals/corr/SimonyanZ14a} convolutional network, since it belongs in the ILSVR challenge 2014 to the best performing models, it represents a popular network structure (convolutional networks), is easily accessible through the Keras \cite{chollet2015keras} framework and has among those within Keras available networks the highest number of parameters (to find out if weight perturbing defense mechanisms are applicable to networks of serious size).

\paragraph{Preprocessing:}
The Keras implementation of the VGG19 network expects its input to be a channel centered (see below) BGR\footnote{
    An RGB image where the blue and the red color channel is swapped.
} image of size $224 \times 224$.
The validation data set of the ILSVRC 2012 did not meet these requirements,
therefore we had to perform the following preprocessing steps to every image:
\begin{enumerate}
    \item convert the image to the BGR color space
    \item uniformly scale the image, such that the smaller side of it has size 224
    \item center-crop the image, such that it is square-shaped afterwards
    \item subtract from every BGR-pixel the overall mean-BGR-pixel of VGG19’s training set: [103.939, 116.779, 123.68]\footnote{
        This value is mentioned here: \url{https://gist.github.com/ksimonyan/3785162f95cd2d5fee77\#file-readme-md}.
    } (this is what we call channel centering)\footnote{
    Note that this is neither the same as subtracting the mean-image (feature-wise centering) nor the same as normalizing each image to have zero-mean (sample-wise centering).
}
\end{enumerate}
Since center-cropping might remove valuable information from the image changing the ground truth, we discard in the next step every image which is not recognized correctly (in terms of top-1 classification) by VGG19. This has also the pleasant side effect of discarding all images not classified correctly by VGG19 in the first place, avoiding to address those examples in later analyses. 35104 images remain after the discarding process, leaving, in our view, still a reasonably sized data set. For the rest of this paper we will refer to this data set as $\mathcal{D}$.

\paragraph{Generating Adversarial Examples:}
Using the framework FoolBox \cite{rauber2017foolbox} (in version 1.6.1) written by Jones Rauber et al.,
we generated adversarial examples based on the data set $\mathcal{D}$ for every adversarial example creating method listed in Table \ref{tab:atk-methods-and-n}.
\begin{table}
    \centering
    \caption{Types and amounts of successfully generated adversarial examples.}
    \begin{tabular}{lr}
        \toprule
        Method & \# of AEs \\
        \midrule
        Fast Gradient Sign Method (\AEMfgsm) \cite{DBLP:journals/corr/GoodfellowSS14} & 35100 \\
        Basic Iterative Method, $\infty$-norm (\AEMbimlinf) \cite{DBLP:journals/corr/KurakinGB16} & 35104 \\
        DeepFool Attack (\AEMdfa) \cite{DBLP:conf/cvpr/Moosavi-Dezfooli16} & 35095 \\
        Saliency Map Attack (\AEMsma) \cite{DBLP:conf/eurosp/PapernotMJFCS16} & 34508 \\
        L-BFGS Attack (\AEMlbfgs) \cite{DBLP:journals/corr/SzegedyZSBEGF13} & 35104 \\
        Carlini Wagner Attack, 2-norm (\AEMcwltwo) \cite{DBLP:conf/sp/Carlini017} & 35104 \\
        \bottomrule
    \end{tabular}
    \label{tab:atk-methods-and-n}
\end{table}

For every attack, the default settings given by FoolBox were used. That means the goal\footnote{%
    The goal is defined by the default FoolBox criterion »Misclassification«: \url{https://foolbox.readthedocs.io/en/latest/modules/criteria.html\#foolbox.criteria.Misclassification}
} of each attack was simply to cause VGG19 to misclassify the generated adversarial example, by means of Top-1 classification and with respect to the ground truth of the corresponding benign image. This is sometimes called untargeted misclassification.

Some of the methods failed to derive adversarial examples from some images in $\mathcal{D}$,
which is why the numbers in the right column of table \ref{tab:atk-methods-and-n} are not always equal to 35104 – the full size of $\mathcal{D}$.

\paragraph{Selecting the Defense Mechanism Parameters}
All considered defense mechanisms are parameterizable, where the parameter is $\sigma$ in the case of the DMs L1, L* and L+ and for RPENN the parameters are $\reldev$ and $\esbsize$. To determine suitable values for $\sigma$, we first tried the values Gu and Rigazio suggest in their paper, namely $\sigma \in \Set{0.01, 0.02, 0.05, 0.1, 0.2, 0.5}$. However, these settings led to poor performance when used for VGG19 on ImageNet (Gu and Rigazio benchmarked small networks on MNIST) against \AEMcwltwo\ adversarial examples. We than tried 50 values linearly evenly spaced across the range 0.01 to 10. This range turned out to give a good overview regarding the quality and efficacy of L1. For L* and L+ however, that range was not suitable either (again poor performance). We ended up using 50 geometrically evenly spaced values from the range $10^{-4}$ to $10^{-1}$, because here the peak of efficacy is in the center of the diagrams, as can be seen in Section \ref{sect:results}. We reused those ranges for the other attack methods.

The data suggested that probing values smaller than the lower bound, or larger than the upper bound of the stated ranges makes no sense, as either the efficacy or the quality decreased considerably. This is plausible when assuming that injecting less and less noise ultimately disables the effect of the defense mechanisms. Conversely, by increasing the magnitude of the injected noise too much, the input’s ground truth might change, or the network’s weights might become defective.

In a brief exploratory search for a suitable range for $\lambda$, we tested the search space $\Set{0.01, 0.02, \dots, 0.24, 0.25, 0.5, 0.75, 1.0, 1.5, 2, 3}$. We found that for $\lambda > 0.5$ the accuracy of \dmnamesing\ on benign input dropped down to non-justifiable level (less than 50\% accuracy), while also not performing well against adversarial input. Therefore we set the upper bound of the range of $\reldev$ to $0.5$ (yet we present in Section \ref{sect:results} only measurements where $\reldev \leq 0.33$ due to layout reasons and poor quality of \dmnamesing\ for $0.33 \leq \lambda \leq 0.5$). A second observation was that the step size of $0.01$ is too subtle, not providing additional insight. We concluded to half the search space even further to save computation time, by using a step size of $0.02$.

We chose the parameter $\esbsize$ to grow exponentially, to cover sufficient ensemble sizes (trying to detect if there’s a limit regarding the effect of further increasing the ensemble size), without consuming too much computation time in contrast to, for example, reaching equally high values for $\esbsize$ by linear growth. The parameter $m$ is always uneven, to reduce the chance of ties if, for example, majority vote is used to summarize the results of the variations.

To identify well performing combinations of $\reldev$ and $\esbsize$, and to evaluate the efficacy of \dmnamesing\ against the various adversarial attacks, we performed an extensive search through product search space of the components
\begin{itemize}
    \item all attack methods mentioned in Table \ref{tab:atk-methods-and-n}
    \item $\lambda \in \Set{0.01, 0.03, 0.05, \dots, 0.47, 0.49}$
    \item $m \in \Set{1, 3, 7, 15, 31, 63, 127}$
\end{itemize}
and recorded two values: $\efficacyvar(\text{\dmnamesing})$ regarding the current attack method and $\qualityvar(\text{\dmnamesing})$ regarding the corresponding\footnote{
    Benign images, where the generation method failed for to create an adversarial image, were omitted.
} benign images.

\paragraph{Estimating Quality, Efficacy and Robustness}
We mentioned that we can only estimate the quality, efficacy and robustness of the studied defense mechanisms, because the randomization they utilize affects these metrics. Regarding $\efficacyvar(d)$ and $\qualityvar(d)$ for $\dmvar \in \Set{\text{L1}, \text{L*}, \text{L+}}$, we recorded $\efficacyvar(d)$ and $\qualityvar(d)$ for 10 different random seeds and averaged the respective measurements. The result is our estimate.
Concerning the estimation of $\efficacyvar(\text{\dmnamesing})$ and $\qualityvar(\text{\dmnamesing})$ for high values of $\esbsize$, we refused repeating each measurement 10 times, with respect to the amount of computational resources needed and also to the self-correcting nature of large ensembles. Thus we repeated the measurements only $\ceil*{\frac{10}{\esbsize}}$ times, leading to generating at least 10 variations per combination, each with a fresh random seed (taking into account the $\esbsize$ variations per combination). In our opinion, the average of the $\ceil*{\frac{10}{\esbsize}}$ measurements still suffices for being a representative estimation.

After having explored which settings lead to a good performance (for details see Section \ref{sect:results}) in terms of quality and efficacy, we selected for each DM two settings which promise to be a good overall choice. One setting guarantees a minimum quality of 99\,\% (among all considered attack methods) and simultaneously achieves a high worst case efficacy (also regarding all considered attack methods). The second setting was selected regarding the same criteria, but for 98\,\% quality.
Table \ref{tab:robusettings} shows settings meeting the mentioned criteria.
\begin{table}
    \centering
    \caption{Defense mechanism parameter settings which lead to a high worst case efficacy among the attack methods in Table \ref{tab:atk-methods-and-n}, still serving a certain minimum quality. The notation $\qualityvar(\dmvar) > x$ means in this specific context the worst case quality of the considered DM among all discussed attack methods in Table \ref{tab:atk-methods-and-n}.}
    \begin{tabular}{lcc}
        \toprule
        $\dmvar$ & $\qualityvar(\dmvar) > 0.99$ & $\qualityvar(\dmvar) > 0.98$ \\
        \midrule
        L1                            & $\sigma = 2.46 $ & $\sigma = 4.09 $ \\
        L*                            & $\sigma = 4.71 \cdot 10^{-4}$ & $\sigma = 8.29 \cdot 10^{-4}$ \\
        L+                            & $\sigma = 4.71 \cdot 10^{-4}$ & $\sigma = 8.29 \cdot 10^{-4}$ \\
        \dmnamesing\ $(\esbsize = 1)$ & $\reldev = 0.03$ & $\reldev = 0.05$ \\
        \dmnamesing                   & $(\reldev, \esbsize) = (0.05, 7)$ & $(\reldev, \esbsize) = (0.11, 63)$ \\
        \bottomrule
    \end{tabular}
    \label{tab:robusettings}
\end{table}
We then recorded the robustness $\robuvar(\dmvar, \robulevel, \ninstalls)$ for $\dmvar \in \Set{\text{L1}, \text{L*}, \text{L+}, \text{\dmnamesing\ } (\esbsize = 1), \text{\dmnamesing}}$, for both configurations of Table \ref{tab:robusettings} respectively, for robustness levels $\robulevel \in \Set{0.5, 0.8, 0.95, 0.99, 1}$ and for $\ninstalls = 128$ (each installation with a fresh random seed). We chose these robustness levels since high values for $\robulevel$ correspond to maximal damage an attacker may cause.

Note that by definition the robustness metric itself sort of is an estimation of the true robustness of a defense mechanism (which could be defined\footnote{We do not in this paper, because we see no benefit by doing so.} to be the expected value over all possible random seeds). Choosing high values for $\ninstalls$ suffices to give a good estimate – further repetition is not necessary.

\section{Experimental Results}
\label{sect:results}
\paragraph{Quality and Efficacy:}
We see in Figure \ref{fig:quali-effi-cwl2-bimlinf-lbfgs} and in Figure \ref{fig:eq-grid-l1starplus} (also in Appendix \ref{app:effi-quali}), that as the magnitude of noise increases (the absolute amount $\sigma$ as well as the relative deviation $\reldev$), the quality of all DMs decreases. The less noise is introduced, the better. To achieve satisfying results for efficacy, one needs to find a sweet spot regarding the amount of noise to inject. If it is too small or too high, the efficacy drops. Additionally, one has to trade-off efficacy with quality, since the peak efficacy values are achieved at quality levels clearly below 100\,\%. On average over all attack methods (and if applicable also over the ensemble sizes), L1 reaches peak efficacy roughly at $94 \pm 2\,\%$ quality; L+/L* reach it roughly at $91 \pm 5\,\%$ quality; \dmnamesing\ without ensemble reaches it roughly at $93 \pm 3\,\%$ and \dmnamesing\ with ensemble reaches it roughly at $96 \pm 3\,\%$. So we can say that high efficacy comes at the price of reduced quality. What loss of quality is acceptable depends, of course, on the situation.
\begin{figure}
    \centering
    \includegraphics[width=0.75\textwidth, height=0.20\textheight]{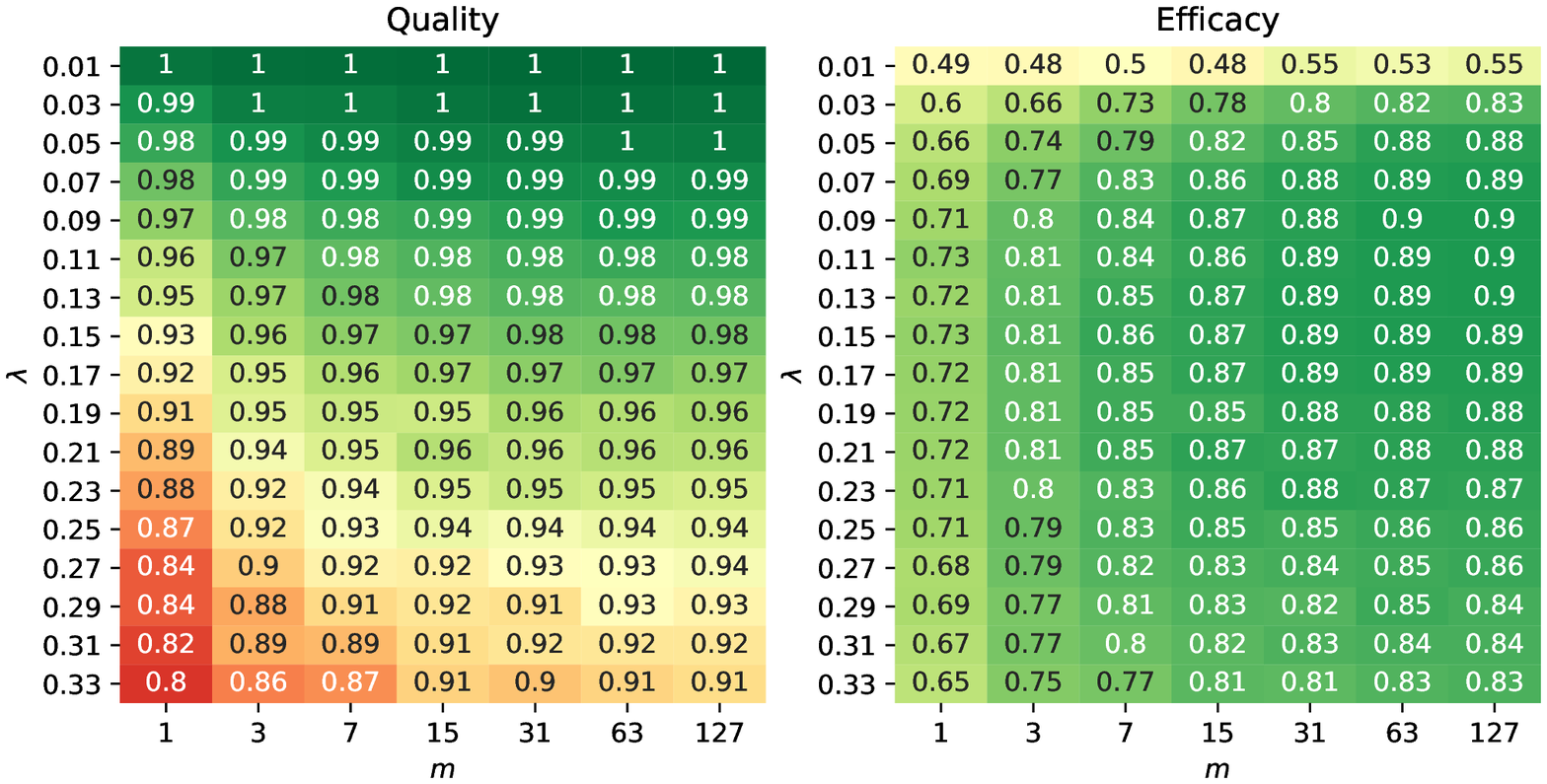}
    \includegraphics[width=0.75\textwidth, height=0.20\textheight]{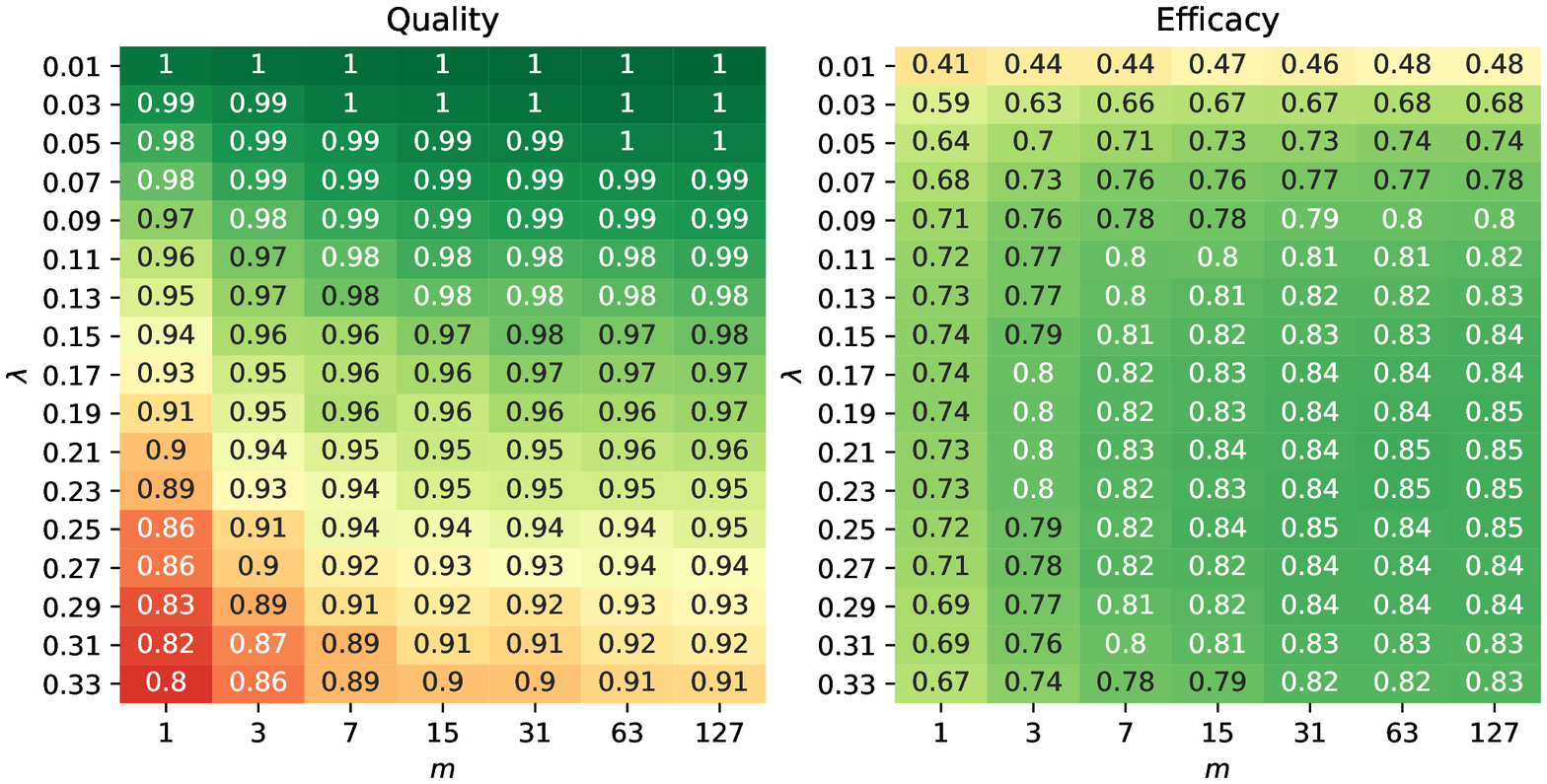}
    \includegraphics[width=0.75\textwidth, height=0.20\textheight]{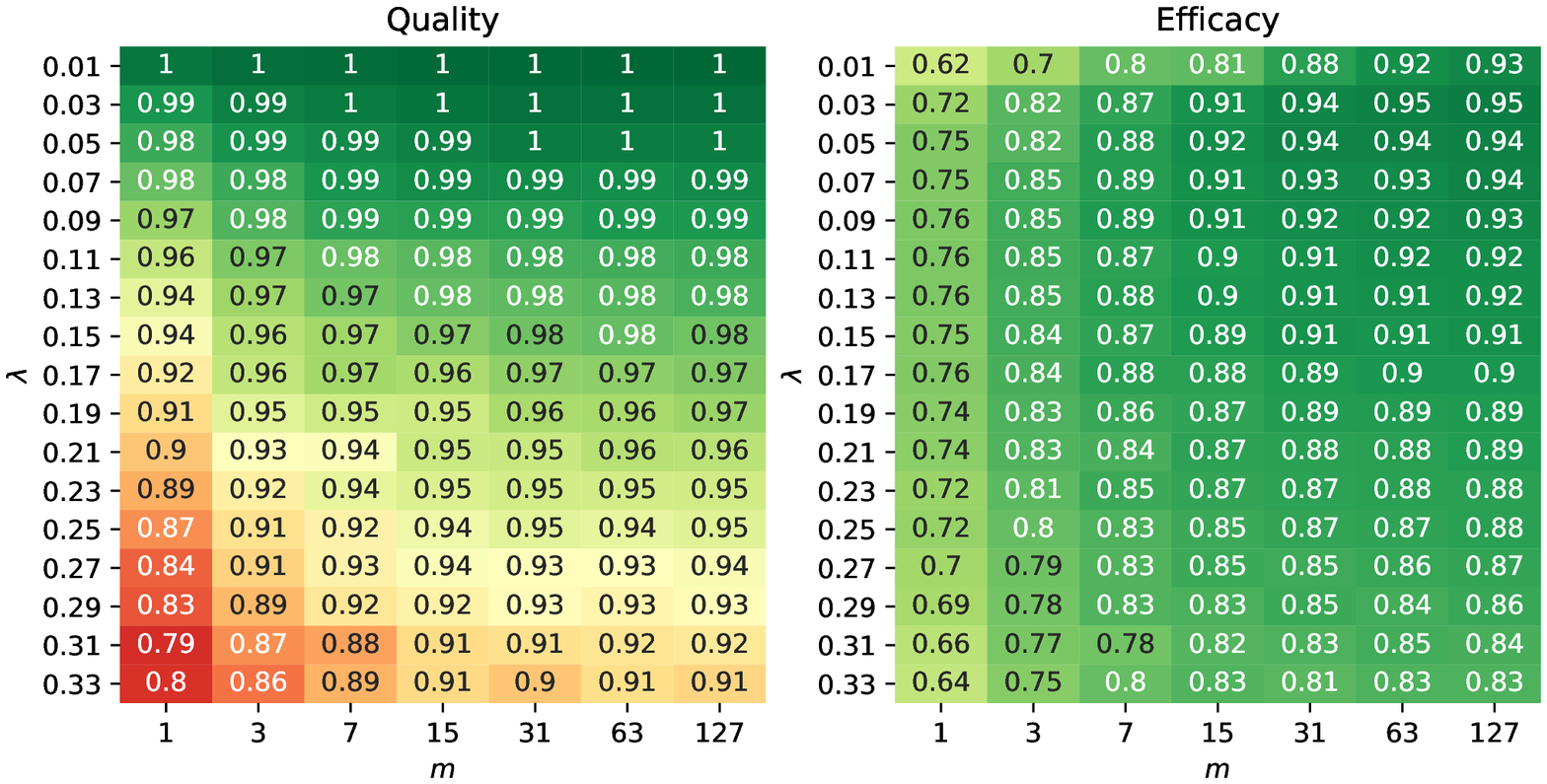}
    \includegraphics[width=0.75\textwidth, height=0.20\textheight]{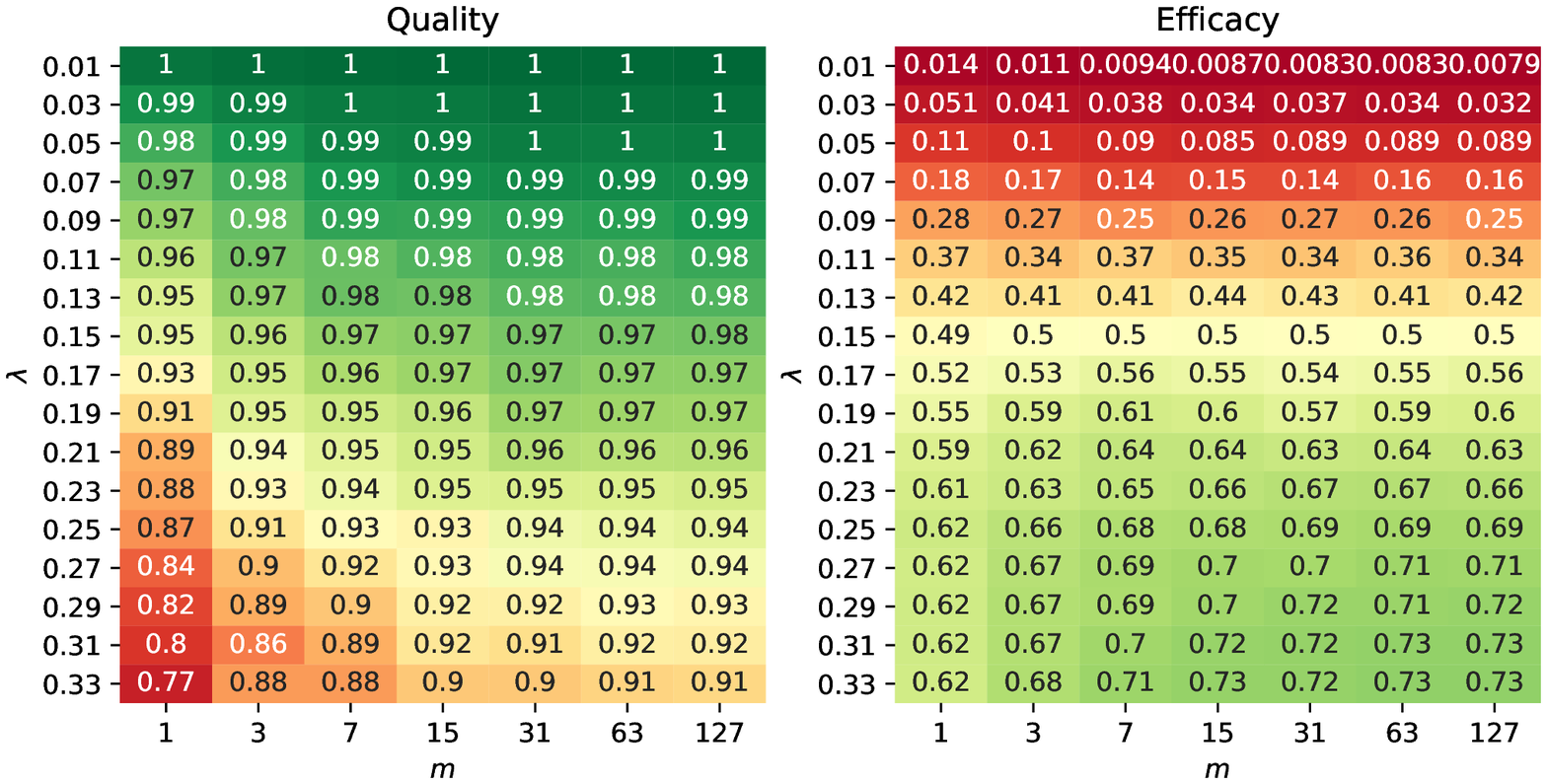}
    \caption{
        The estimated values of $\qualityvar(\text{\dmnamesing})$ (left) and $\efficacyvar(\text{\dmnamesing})$ (right) using various settings for $\reldev$ and $\esbsize$ against \AEMdfa\ (top row), \AEMcwltwo\ (second row), \AEMbimlinf\ (third row) and \AEMlbfgs\ (bottom row).
        Note that the quality and efficacy values shown in the cells of the tables are rounded to 3 decimal places. A value of $1$ just means an accuracy of greater than 99.5\,\%. None of the shown parameter settings lead to perfect quality.
    }
    \label{fig:quali-effi-cwl2-bimlinf-lbfgs}
\end{figure}
\begin{figure*}
    \centering
    \includegraphics[width=0.75\textwidth]{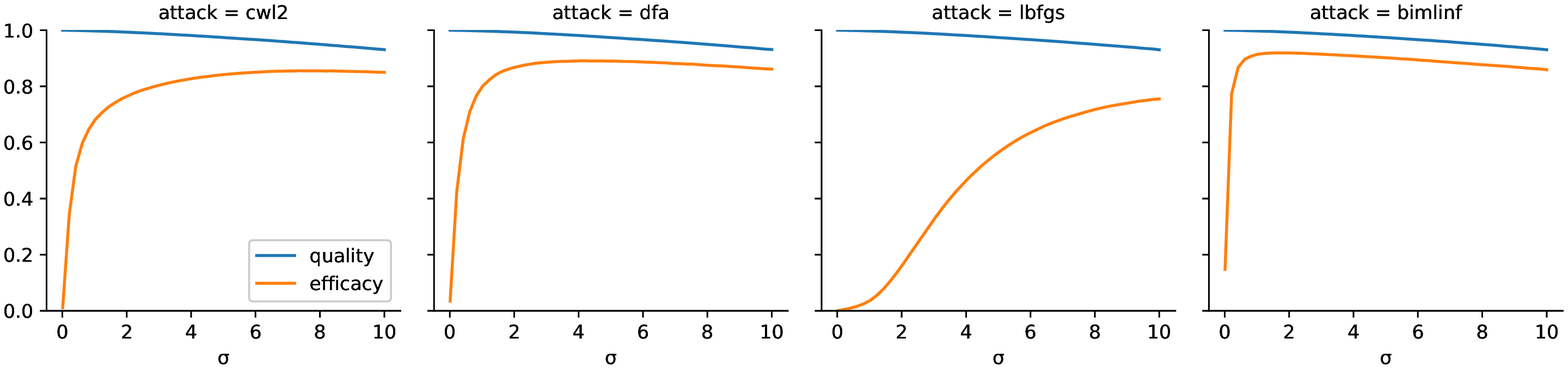}
    \includegraphics[width=0.75\textwidth]{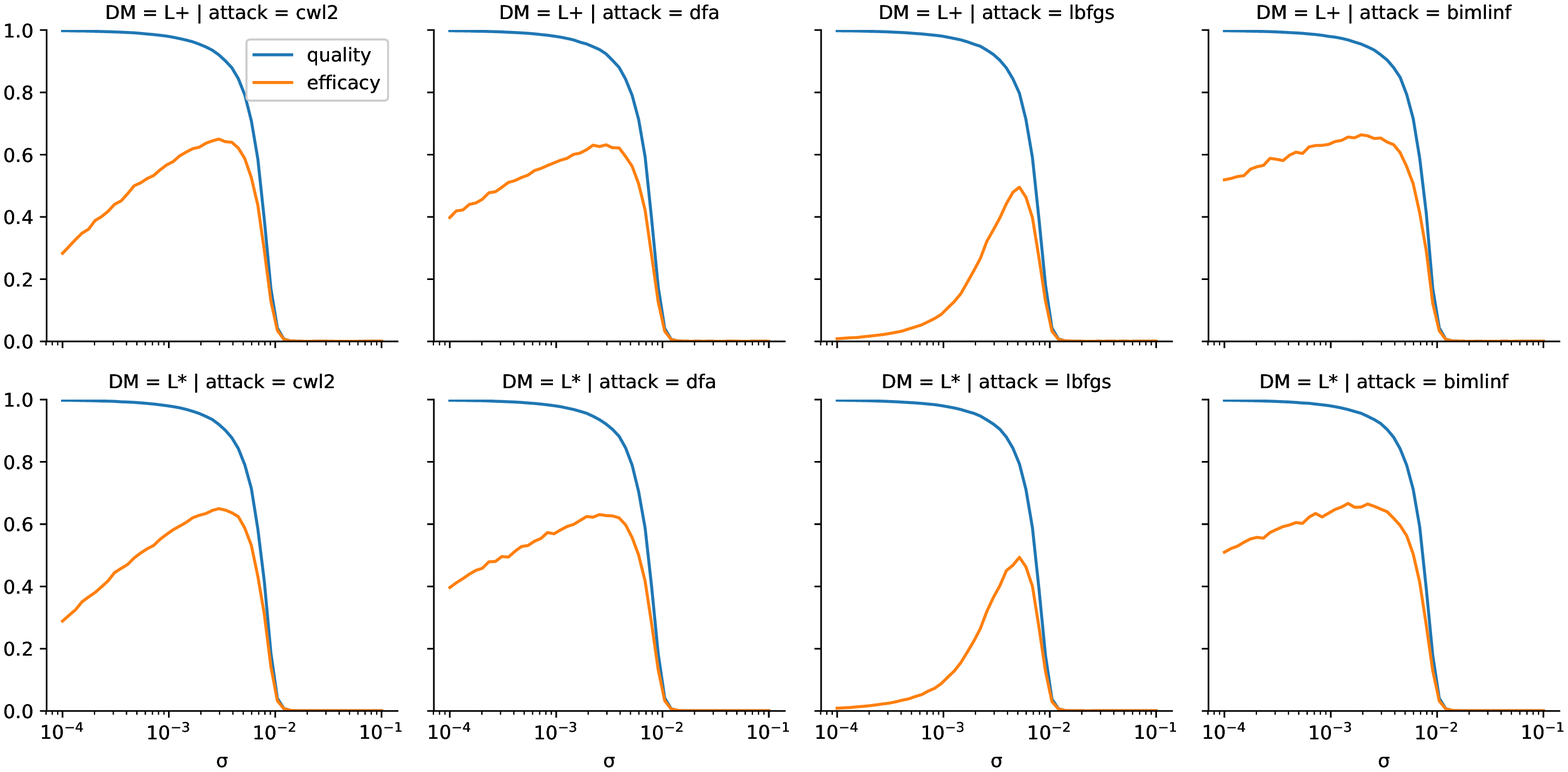}
    \caption{
        The quality and efficacy of L1 (first row) and L*/L+ (second/third row) against the attack methods \AEMcwltwo, \AEMdfa, \AEMlbfgs\ and \AEMbimlinf\ (columns). Notice the linear scale on the first row’s x-axes, whereas the other two row’s x-axes scale logarithmically.
    }
    \label{fig:eq-grid-l1starplus}
\end{figure*}

\paragraph{Robustness:}
In Figure \ref{fig:robu-grid-99} (and also in appendix \ref{app:robu}), where the robustness of the defense mechanisms against various attacks is shown, we see that as the number of installations $\ninstalls$ grows, the proportion of AEs which successfully fool most or all out of of the $\ninstalls$ installations decreases clearly. This indicates that randomness has a positive influence on this metric. For a non-random DM it would make no difference to attack 1 out of 1 or $\ninstalls$ out of $\ninstalls$ – an adversarial example which overcomes on installation would overcome all of them.

Comparing the Figures \ref{fig:robu-grid-99} and \ref{fig:robu-grid-99-app} directly with Figure \ref{fig:robu-grid-98}, we see that we generally accomplish only a minor gain in terms of additional robustness, if we trade in quality for more efficacy by selecting DM settings from the right column of Table \ref{tab:robusettings}. Against most attacks (namely \AEMdfa, \AEMbimlinf, \AEMfgsm, \AEMsma\ and to some extent also \AEMcwltwo), the more conservative settings from the left column of Table \ref{tab:robusettings} suffice to attain desirable robustness. Only regarding the attack \AEMlbfgs, or robustness levels $\robulevel \leq 0.5$ we think it’s worth considering less conservative settings to achieve more robustness.
\begin{figure*}
    \centering
    \includegraphics[width=0.75\textwidth]{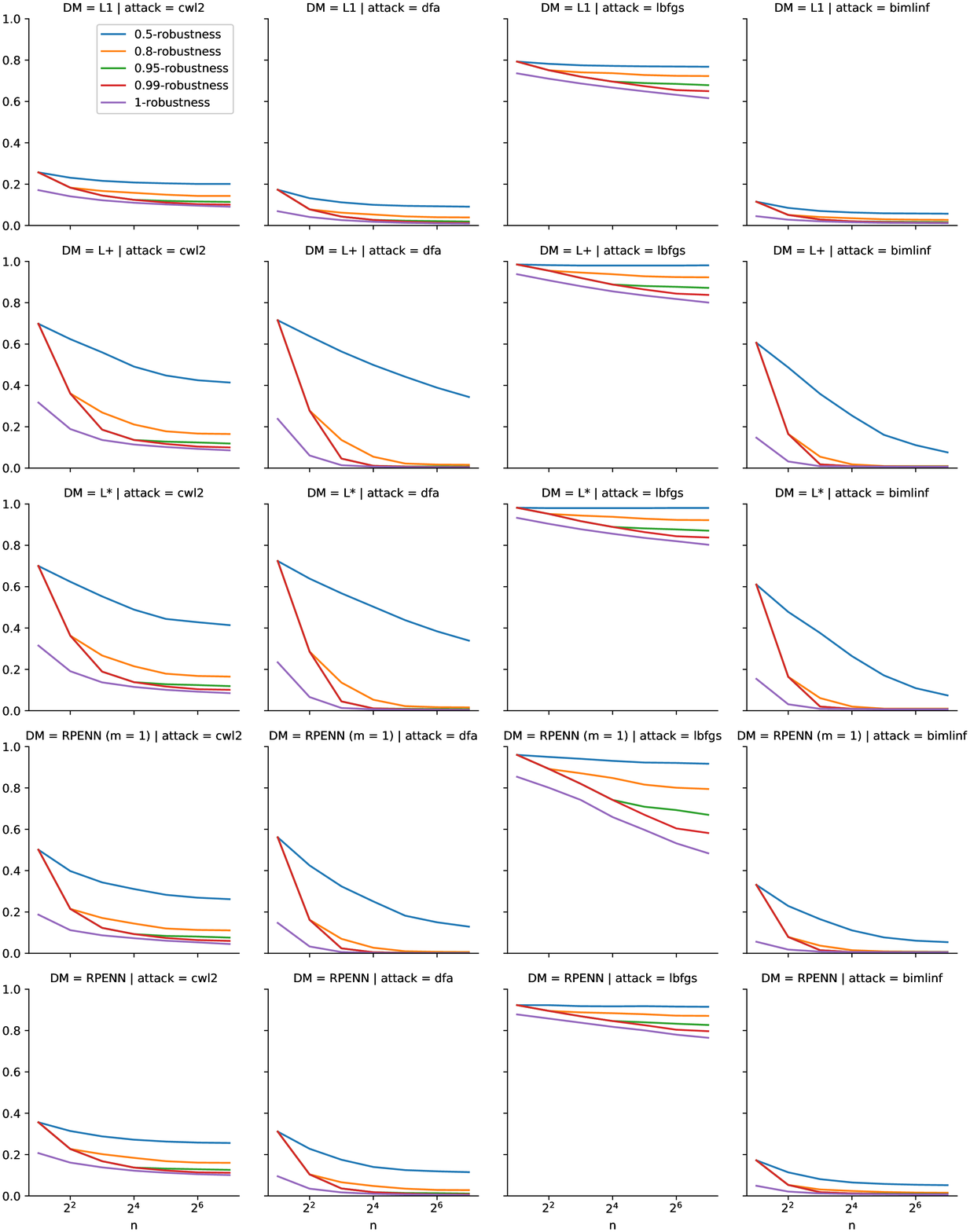}
    \caption{
        The values of $\robuvar(\dmvar, \robulevel, 128)$, for $(\dmvar, \robulevel) \in \Set{\text{L1}, \text{L*}, \text{L+}, \text{\dmnamesing\ } (m = 1), \text{\dmnamesing}} \times \Set{0.50, 0.80, 0.95, 0.99, 1.00}$, of all five defense mechanisms (rows) against the attack methods \AEMcwltwo, \AEMdfa, \AEMlbfgs\ and \AEMbimlinf (columns).
        The DMs are configured with the settings listed in the column of Table \ref{tab:robusettings} which is labeled with $\qualityvar(\dmvar) > 0.99$.
        To see a trend, intermediate values of $\robuvar(\dmvar, \robulevel, \ninstalls)$ for $\ninstalls \in \Set{2, 4, \dots, 64}$ are plotted too.
    }
    \label{fig:robu-grid-99}
\end{figure*}

Imagining the number of installations one might expect when a DNN is shipped via an App to smartphones, or as part of a cyber-physical system to autonomous cars, we rate the values for $\ninstalls$, where one already can see reasonable results, to be surprisingly low. However, since not all robustness values reliably tend to 0, we expect there exist adversarial images, which seem to be quite immune to the randomization introduced by the evaluation defense mechanisms.

\paragraph{Efficiency:}
In addition to analyzing DMs on quality, efficacy, and robustness, we also considered whether they are efficient. Informally, we expect good defense mechanisms to only add a constant overhead, where this constant is very small value. From this perspective, all the DMs seem very good. More specifically, the resource consumption of L1 (both time and space) scales linear with the input dimension, which is a constant. Hence, L1 introduces just constant overhead and a small one at that. Both L+ and L* also introduce only a constant overhead. The overhead of \dmnamesing\ scales with network and ensemble size. However, given that the network size is fixed for a given DNN and the ensemble size can also be fixed ahead of time, once again the overhead is a constant. We didn't perform detailed empirical analysis of the efficiency of these DMs, but our experience with them suggests that the actual overhead we see in practice is a small constant. While \dmnamesing\ is slower than L1, L+, and L*, its running time is still tolerable for small to medium ensemble sizes. For large ensemble sizes one has to use a more sophisticated implementation, where the variations of the ensemble are not generated serially, like they are created in our reference implementation.

\paragraph{General Results:}
Surprisingly, the simplest of the considered defense mechanisms, namely L1, seems according to our metrics to be the best overall defense mechanism. Only \dmnamesing\ with a large ensemble accomplishes similar efficacy and quality, but not as good robustness and clearly worse efficiency. \dmnamesing\ without ensemble is generally on par, or superior to L+ and L* and shows a promising robustness behavior (even against \AEMlbfgs).
The similar performance of L* and L+ throughout all metrics we attribute to the fact, that the noise injected into the input and into the hidden layers is controlled by the same parameter $\sigma$. We expect to see different behavior of L*, if the noise injected to the hidden layer is controlled by another (and usually higher valued) parameter.

According to our metrics, the \AEMlbfgs\ attack method is the strongest among the attacks (all considered DMs perform significantly worse against it than against all other attacks). The second strongest attack is the \AEMcwltwo\ attack. The weakest attack is the \AEMbimlinf attack.

\section{Related Work}
\label{sect:related}
Jin, Dundar and Culurciello proposed in \cite{DBLP:journals/corr/JinDC15} a training method called \emph{stochastic feedforward}, involving randomized injection of Gaussian noise into the input layer and specific types of hidden layers during the training process. On this point, they differ from the presented DMs, which are applied to already trained networks (and are therefore easier to deploy).

Liu et al. \cite{DBLP:journals/corr/abs-1712-00673} propose a technique they call \emph{Random Self-Ensemble}, where they combine randomness and ensembles (like \dmnamesing). They insert {\it noise layers} into feed-forward networks (which is like injecting noise into the network activations – not into the network weights) and retrain it using a special procedure.

Perhaps the defense mechanisms that are closest to ours, are the ones proposed by Gu and Rigazio~\cite{DBLP:journals/corr/GuR14}, which we covered extensively in Section \ref{sect:dm} and \ref{sect:results}. The reason we evaluate Gu and Rigazio’s work, but not Jin, Dundar and Culurciello’s work, as well as not the work of Liu et al., is deployability. For this paper we focused on easily deployable DMs. Defense mechanisms which require retraining, for example, are out of this paper’s scope.

Carlini and Wagner propose in \cite{DBLP:conf/ccs/Carlini017} a methodology regarding how to evaluate defense mechanisms against adversarial examples. We incorporate their work in this paper and extend it regarding additional metrics.
\section{Potential Threats to Validity}
\label{sect:validity}
In this section, we examine several potential sources of weakness in the experimental evaluation we conducted, following recommendation by the SIGPLAN committee on such experiments~\cite{SIGPLAN-checklist}, and discuss ways in which we have attempted to minimize the impact of these factors.

\paragraph{Chosen Data Set:} The choice of data set can have significant impact on empirical evaluation of DMs. We protect against this possible weakness by choosing a well-known, large, and comprehensive data set for our experiments: the ILSVRC2012 validation data set. We believe that, as a consequence, our results are quite robust.

\paragraph{Attacker Model:} Following the taxonomy given by Papernot et al. \cite{DBLP:journals/corr/PapernotMSW16}, we define a fairly precise attacker model. The attacker we consider is powerful and has white-box access to the DNN under attack. The only limitation we place is that the attacker may not a priori know the random seed used to randomize the considered DMs.

\paragraph{Experimental Methodology:} We follow the experimental methodology suggested by Carlini and Wagner \cite{DBLP:conf/ccs/Carlini017} for all our experimental evaluation, including using the ImageNet data set as suggested by them. We further extend their methodology by defining a set of properties and metrics for quality, efficacy, and robustness to characterize and differentiate good defense mechanisms from bad ones.

\paragraph{Attacks Methods Considered:} We tested our defense mechanism thoroughly over six different and common gradient-based attack methods. We show that for all but one of these attack methods (\AEMlbfgs\ is the exception) the L1 and the \dmnamesing\ defense mechanism perform reasonably well according to our metrics. Having said that, it is possible that non gradient-based attack methods or black-box attack methods (for example the method Papernot et al. suggest in \cite{DBLP:conf/ccs/PapernotMGJCS17}) succeed against these defense mechanisms. 
\section{Conclusions}
\label{sect:conc}
Adversarial attacks are a very important problem for machine learning in general. It is critical that this problem be appropriately addressed to ensure continued adoption of reliable and secure machine learning-based systems. In response to this problem, many researchers have proposed defense mechanisms, both randomized and otherwise. However, to-date a robust methodology to evaluate these DMs has been lacking. In our work, we proposed a comprehensive methodology and a set of properties and metrics to evaluate DMs. An important part of our methodology is evaluating the DMs on a real world size DNN (VGG19) and on a large and complex data set (ImageNet). While this part seems obvious, so far it has not been adhered to by many researchers. Second, it is critical that the DMs be tested against a variety of attack methods, including strong attacks like \AEMlbfgs\ and \AEMcwltwo. Once again, prior to our work we have not yet come across such robust testing. Finally, it is critical to define a set of properties against which to evaluate and test DMs. We defined 4 such properties, namely, efficacy, quality, robustness, and efficiency. (As this field develops further, we expect more properties will be added to our list.)

Using the above-mentioned methodology, we tested 4 different DMs, namely, L1, L*, L+, and RPENN. Two of these DMs (L+ and \dmnamesing) are our novel contribution. Surprisingly and counter-intuitively, our evaluation showed that L1 (where only the input is perturbed) performs the best in terms of overall efficacy, quality, robustness, and efficiency. We expected that the perturbation of the defended DNN and the usage of an ensemble would yield superior results.

\bibliographystyle{plainnat}
\bibliography{bibliography}

\appendix
\section{Additional Data on Efficacy and Quality}
\begin{figure}[h!]
    \centering
    \includegraphics[width=0.75\textwidth,height=0.30\textheight]{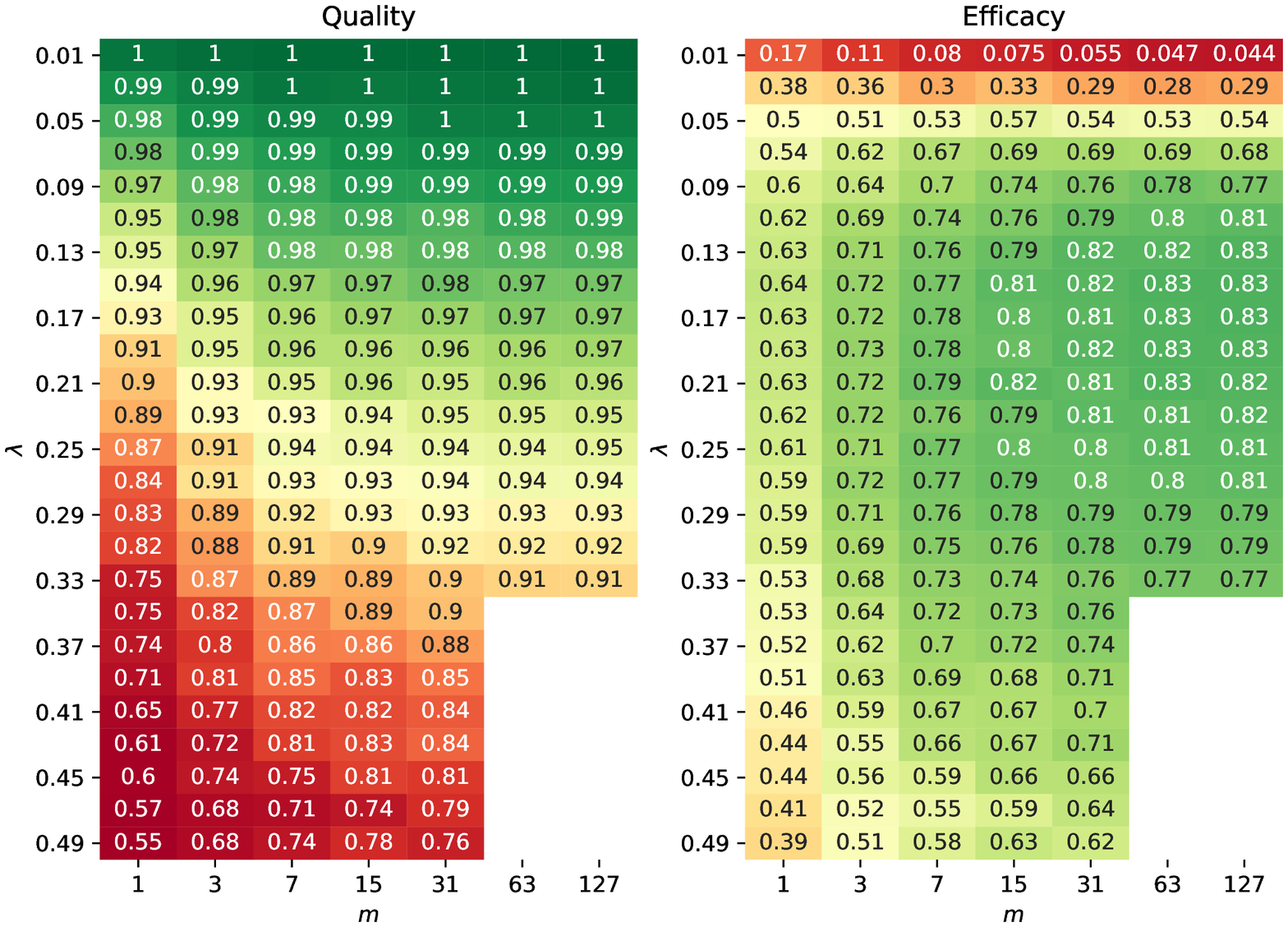}
    \includegraphics[width=0.75\textwidth,height=0.20\textheight]{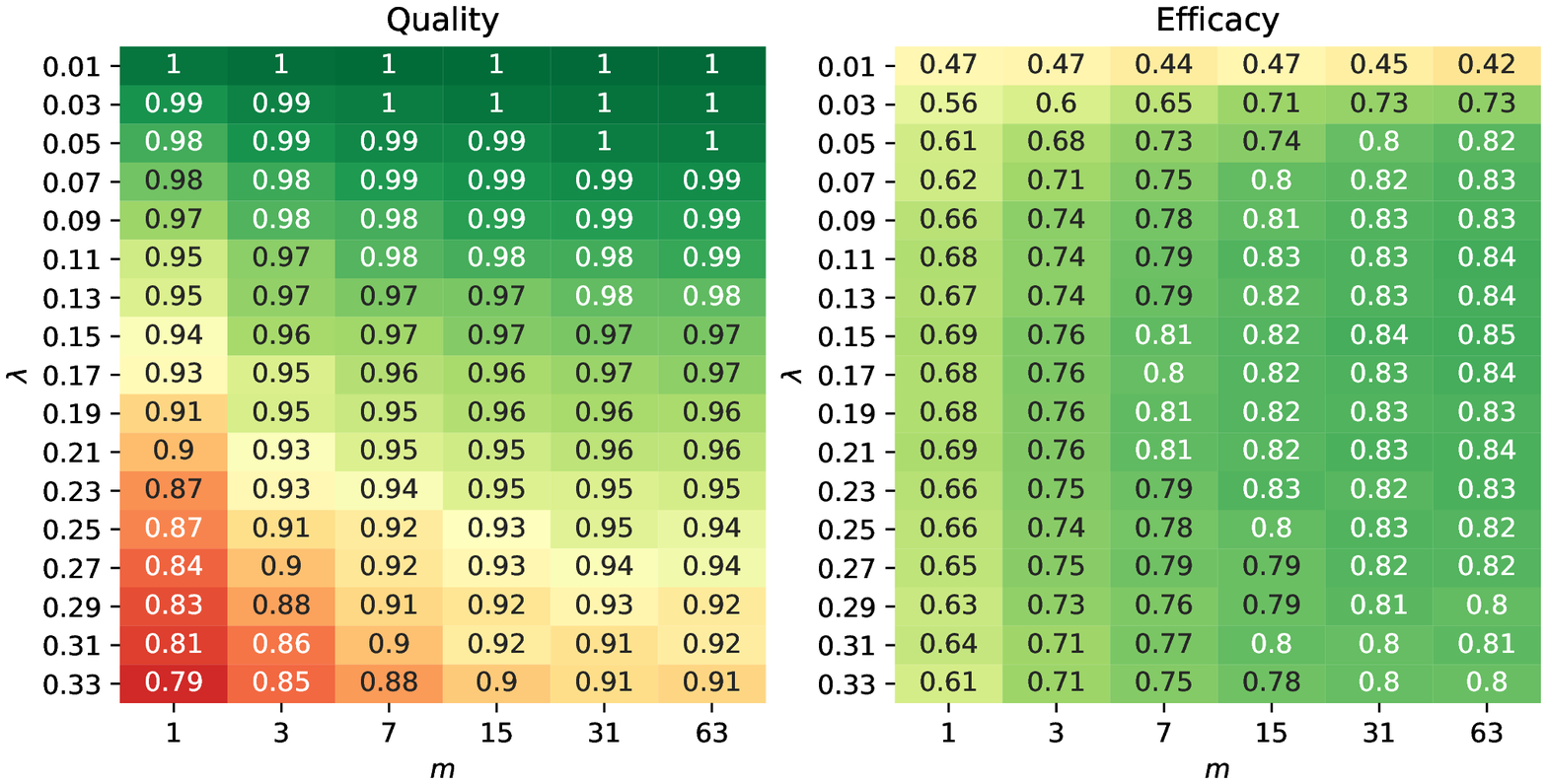}
    \caption{
        Estimated $\qualityvar(\text{\dmnamesing})$ (left) and $\efficacyvar(\text{\dmnamesing})$ (right) using various settings for $\reldev$ and $\esbsize$. The attack methods are \AEMfgsm\ (top) and \AEMsma\ (bottom).
        We saw no reason for computing $\efficacyvar$ and $\qualityvar$ for $\reldev > 0.33$ and $\esbsize \in \Set{63, 127}$, since the data suggests to expect low quality and decreasing efficacy in that region.
    }
    \label{fig:quali-effi-fgsm-ga}
\end{figure}
\label{app:effi-quali}
\begin{figure}[h!]
    \centering
    \includegraphics[width=0.75\textwidth]{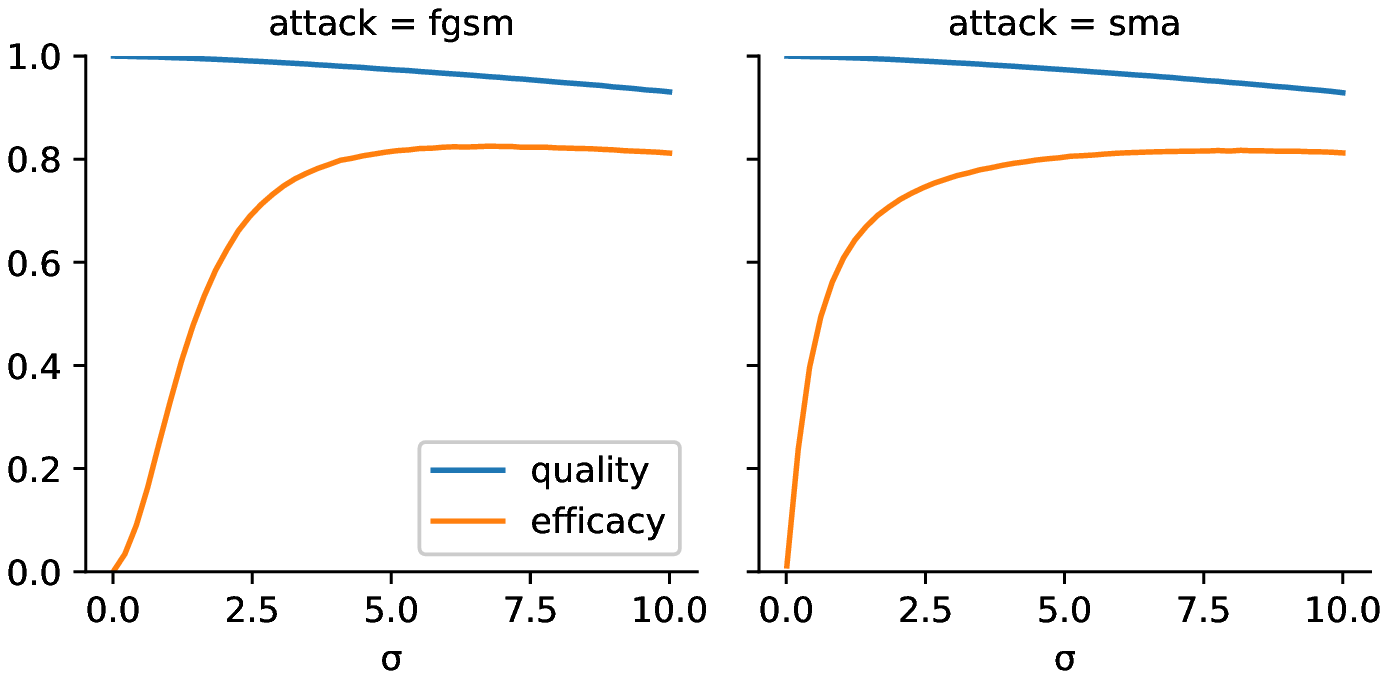}
    \includegraphics[width=0.75\textwidth]{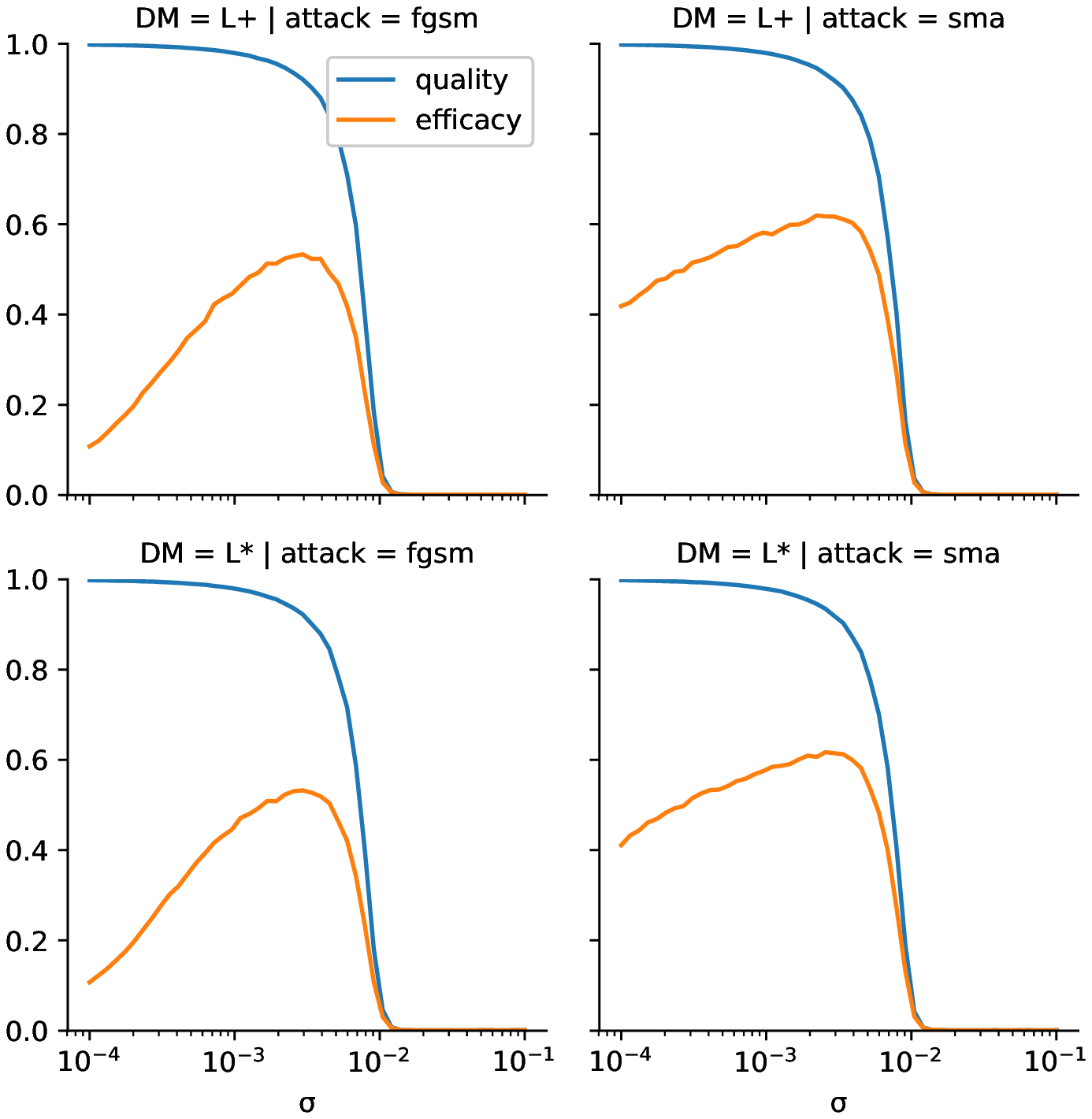}
    \caption{
        The quality and efficacy of L1 (first row) and L*/L+ (second/third row) against the attack methods \AEMfgsm\ and \AEMsma\ (columns). Notice the linear scale on the first row’s x-axes, whereas the other two row’s x-axes scale logarithmically.
    }
    \label{fig:eq-grid-l1starplus-app}
\end{figure}

\section{Additional Data on Robustness}
\label{app:robu}
\begin{figure}[h!]
    \centering
    \includegraphics[height=0.75\textheight]{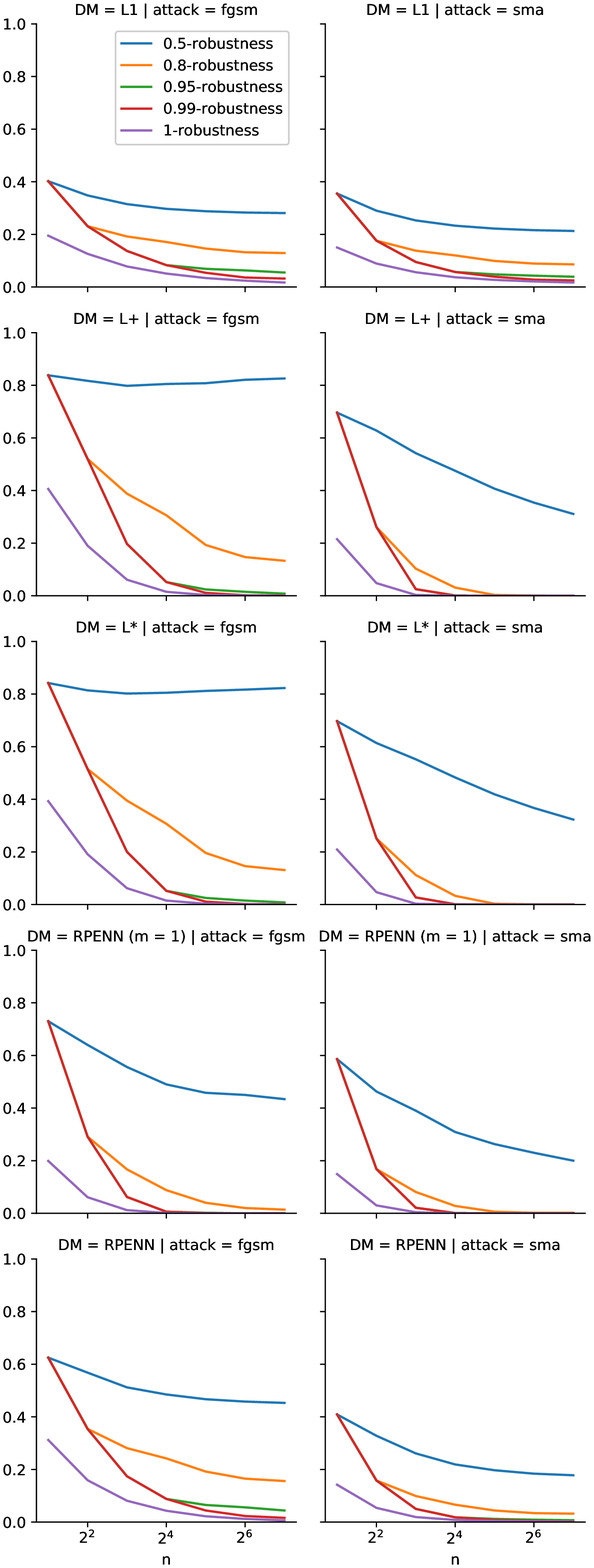}
    \caption{
        The values of $\robuvar(\dmvar, \robulevel, 128)$, for $(\dmvar, \robulevel) \in \Set{\text{L1}, \text{L*}, \text{L+}, \text{\dmnamesing\ } (m = 1), \text{\dmnamesing}} \times \Set{0.50, 0.80, 0.95, 0.99, 1.00}$, of all five defense mechanisms (rows) against the attack methods \AEMfgsm\ and \AEMsma\ (columns).
        The DMs are configured with the settings listed in the column of Table \ref{tab:robusettings} which is labeled with $\qualityvar(\dmvar) > 0.99$.
        To see a trend, intermediate values of $\robuvar(\dmvar, \robulevel, \ninstalls)$ for $\ninstalls \in \Set{2, 4, \dots, 64}$ are plotted too.
    }
    \label{fig:robu-grid-99-app}
\end{figure}

\begin{figure*}[h!]
    \centering
    \includegraphics[width=\textwidth]{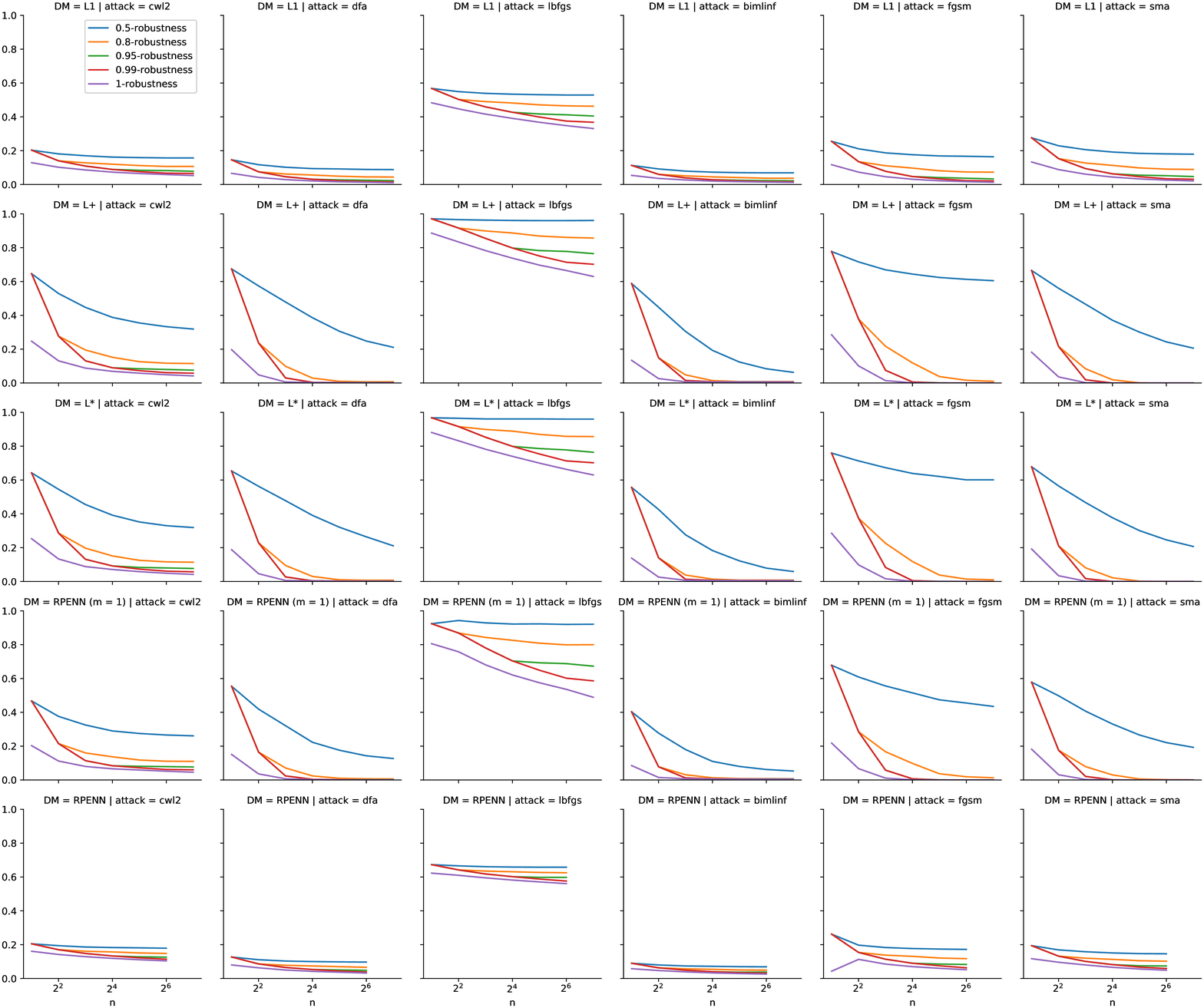}
    \caption{
        The values of $\robuvar(\dmvar, \robulevel, 128)$, for $(\dmvar, \robulevel) \in \Set{\text{L1}, \text{L*}, \text{L+}, \text{\dmnamesing\ } (m = 1), \text{\dmnamesing}} \times \Set{0.50, 0.80, 0.95, 0.99, 1.00}$, of all five defense mechanisms (rows) against the attack methods \AEMcwltwo, \AEMdfa, \AEMlbfgs, \AEMbimlinf, \AEMfgsm\ and \AEMsma\ (columns).
        The DMs are configured with the settings listed in the column of Table \ref{tab:robusettings} which is labeled with $\qualityvar(\dmvar) > 0.98$.
        To see a trend, intermediate values of $\robuvar(\dmvar, \robulevel, \ninstalls)$ for $\ninstalls \in \Set{2, 4, \dots, 64}$ are plotted too.
        Note that, due to time limitations, the robustness values for \dmnamesing\ are only computed up to $n = 64$ instead of $n = 128$.
    }
    \label{fig:robu-grid-98}
\end{figure*}

\end{document}